\documentclass[letterpaper, 10 pt, conference]{ieeeconf}  

\IEEEoverridecommandlockouts                              

\usepackage{latexsym}
\usepackage{times}
\usepackage{epsfig}
\usepackage{graphicx}
\usepackage{amsmath}
\usepackage{amssymb}

\usepackage{multirow}
\usepackage{booktabs}
\usepackage{pifont}
\usepackage{lipsum}
\usepackage{subfigure}

\usepackage{color, colortbl}

\usepackage[pagebackref=true,breaklinks=true,letterpaper=true,colorlinks,bookmarks=false]{hyperref}

\usepackage{color, colortbl}
\newif\ificcvfinal
\iccvfinalfalse
\def\iccvfinalcopy{\global\iccvfinaltrue}

\iccvfinalcopy

\newcommand\paraspace{\vspace*{0.5ex}}
\providecommand\parab[1]{\paraspace\noindent\textbf{#1}}

\newcommand{\txbu}[1]{\underline{\textbf{#1}}}

\newcommand{\dbname}[0]{\ificcvfinal WOMD-LiDAR\else Anonymous-LiDAR\fi}
\newcommand{\wf}[0]{WayFormer}
\newcommand{\swf}[0]{SWFormer}
\newcommand{\lidar}[0]{LiDAR}

\newcommand{\cmark}{\ding{51}}%

\newcommand{\figref}[1]{Fig.~\ref{#1}}

\newcommand{\revonly}[1]{\ificcvfinal\else{#1}\fi}
\newcommand{\arxivonly}[1]{\ificcvfinal{#1}\else\fi}

\title{\bf\dbname: Raw Sensor Dataset Benchmark for\\Motion Forecasting}

\ificcvfinal

\author{Kan Chen, Runzhou Ge, Hang Qiu, Rami AI-Rfou, Charles Qi, Xuanyu Zhou, Zoey Yang, Scott Ettinger, \\ Pei Sun, Zhaoqi Leng, Mustafa Baniodeh, Ivan Bogun, Weiyue Wang, Mingxing Tan, Dragomir Anguelov
\thanks{*This work was done in Waymo LLC}
}

\else

\author{First Author\\
Institution1\\
Institution1 address\\
{\tt\small firstauthor@i1.org}
\and
Second Author\\
Institution2\\
First line of institution2 address\\
{\tt\small secondauthor@i2.org}
}
\fi

\usepackage{etoolbox}
\usepackage[singlelinecheck=false]{caption}
\makeatletter
\patchcmd{\@makecaption}
  {\scshape}
  {}
  {}
  {}
\makeatletter
\patchcmd{\@makecaption}
  {\\}
  {:\ }
  {}
  {}
\makeatother

\begin{document}

\maketitle
\thispagestyle{empty}
\pagestyle{empty}

\begin{abstract}
Widely adopted motion forecasting datasets substitute the observed sensory inputs with higher-level abstractions such as 3D boxes and polylines.
These sparse shapes are inferred through annotating the original scenes with perception systems' predictions.
Such intermediate representations tie the quality of the motion forecasting models to the performance of computer vision models.
Moreover, the human-designed explicit interfaces between perception and motion forecasting typically pass only a subset of the semantic information present in the original sensory input.
To study the effect of these modular approaches, design new paradigms that mitigate these limitations, and accelerate the development of end-to-end motion forecasting models,
\revonly{we release a new large-scale, high-quality, diverse {\lidar} dataset for the motion forecasting task.}
\arxivonly{we augment the Waymo Open Motion Dataset (WOMD) with large-scale, high-quality, diverse {\lidar} data for the motion forecasting task.}

\revonly{Our new dataset (\dbname) consists of over 100,000 scenes that each spans 20 seconds,}
\arxivonly{The new augmented dataset (\dbname)\arxivonly{\footnote{\url{https://waymo.com/open/data/motion/}}} consists of over 100,000 scenes that each spans 20 seconds,}
consisting of well-synchronized and calibrated high quality {\lidar} point clouds captured across a range of urban and suburban geographies.
Compared to Waymo Open Dataset (WOD), {\dbname} dataset contains 100$\times$ more scenes.
Furthermore, we integrate the {\lidar} data into the motion forecasting model training and provide a strong baseline.
Experiments show that the {\lidar} data brings improvement in the motion forecasting task.
We hope that {\dbname} will provide new opportunities for boosting end-to-end motion forecasting models.

\end{abstract}


\section{Introduction}

Motion forecasting plays an important role for planning in autonomous driving systems and received increasing attention in the research community~\cite{chai2019multipath,gao2020vectornet,rhinehart2019precog,tolstaya2021identifying,zhao2021tnt,tang2019multiple}.
The prohibitively expensive storage requirements for publishing raw sensor data for driving scenes limited the major motion forecasting datasets~\cite{ettinger2021large,Argoverse2,caesar2020nuscenes,Woven2020dataset,zhan2019interaction}. 
They instead release abstract representations, such as 3D boxes from pre-trained perception models (for objects) and polylines (for maps), to represent the driving scenes.

\begin{figure}[t]
\centering
\subfigure[Sophisticated interactions with (left) and without {\lidar} (right).]
{
\centering
  \includegraphics[width=0.49\linewidth]{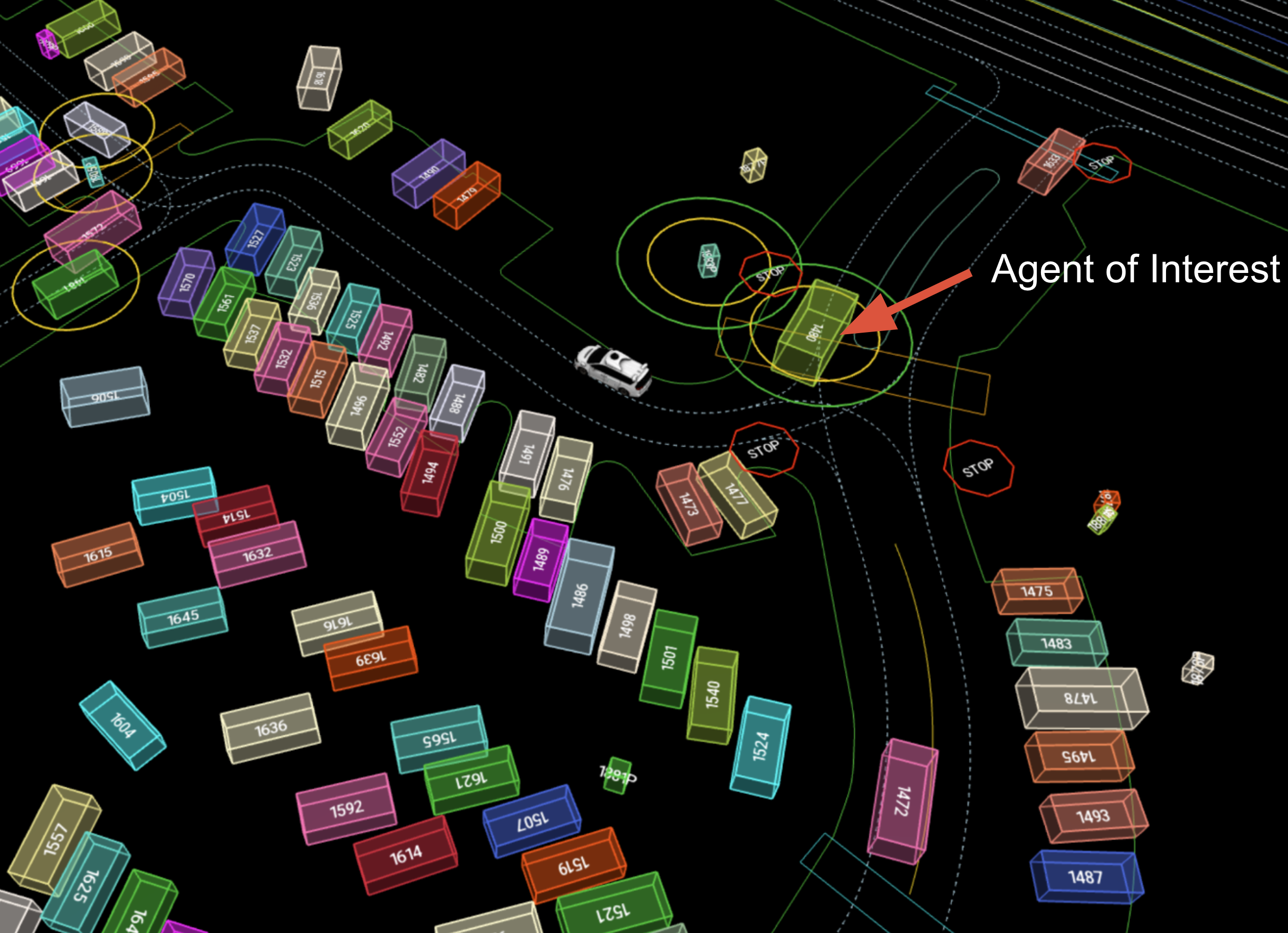}
  \hfill
     \includegraphics[width=0.49\linewidth]{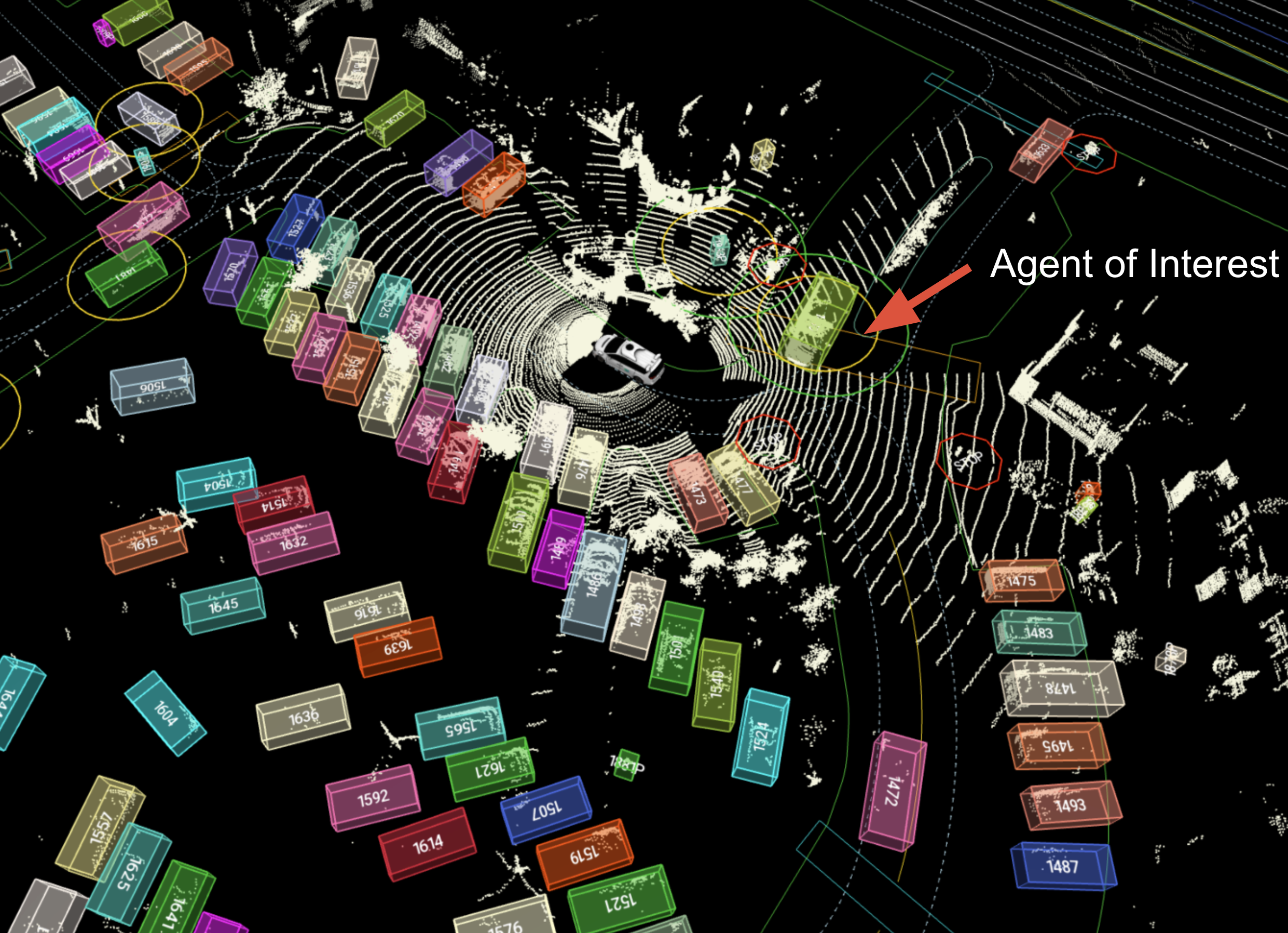}
}
\vspace{-3mm}
\subfigure[Predicted trajectories with (left) and without LiDAR data (right).]{
   \includegraphics[width=\linewidth]{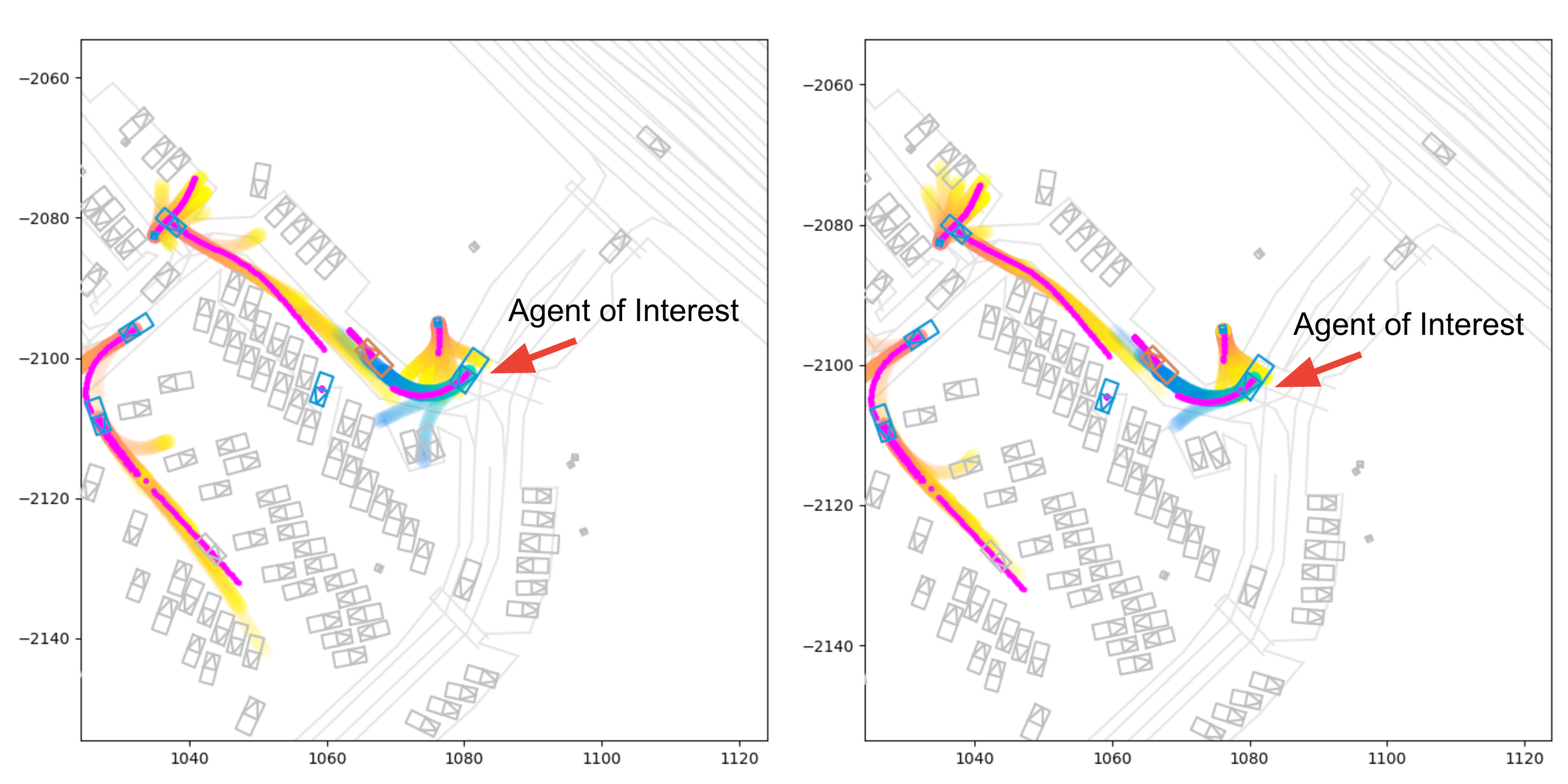}
}

\caption{Human-interpretable labels from the perception system provide limited information at the scene level and the object level. In sophisticated scenes with interaction between multiple objects, raw sensor data provides rich information and helps improve the motion forecasting performance. Legends in the figure: Yellow and blue (highlighted) trajectories are predictions for different agents. Red dotted lines are  agents' ground truth trajectories.}
\label{fig:intro}
\vspace{-6mm}
\end{figure}
\begin{table*}
\definecolor{LightCyan}{rgb}{0.88,1,1}
\begin{center}
\small
\begin{tabular}{c | *{6}{c}}
\toprule
& INTERACTION  & Woven Planet & Shifts  & Argoverse 2 & nuScenes & {\dbname} \\
\midrule
\rowcolor{LightCyan}
Has {\lidar} Data &  &  &  &  & \cmark & \cmark \\
\# Segments & - & 170k & 600k & 250k & 1k & 104k \\
Segment Duration & - & 25s & 10s  & 11s & 20s & 20s  \\
Total Time & 16.5h & 1118h & 1667h  & 763h & 5.5h & 574h \\
Unique Roadways & 2km & 10km & - & 2220km & - & 1750km \\
Sampling Rate & 10Hz & 10Hz & 5Hz & 10Hz & 2Hz  & 10Hz  \\
\# Cities  & 6 & 1 & 6 & 6  & 2 & 6 \\
3D Maps & & & \cmark & \cmark  & & \cmark \\
Dataset Size\textsuperscript{\textdagger} & - & 22GB  & 120GB & 58GB & 48GB & 2.29TB* \\
\bottomrule
\end{tabular}
\end{center}
\caption{Comparison of the popular behavior prediction and motion forecasting datasets. We compare our {\dbname} with INTERACTION~\cite{zhan2019interaction}, Woven Planet~\cite{Woven2020dataset}, Shifts~\cite{malinin2021shifts}, Argoverse 2~\cite{Argoverse2}, nuScenes~\cite{caesar2020nuscenes}. ``-'' indicates that the data is not available or not applicable.
{\textdagger}The sizes are cited from~\cite{Argoverse2}.
*{\dbname} dataset size is after  $\sim8\times$ compression.
}
\vspace{-3mm}
\label{tab:stat_db_intro}
\end{table*}

The absence of the raw sensor data leads to the following limitations:
1) Motion forecasting relies on lossy representation of the driving scenes  (\figref{fig:intro}).
The human designed interfaces lack the specificity required by the motion forecasting task.
For example, the taxonomy of the agent types in Waymo Open Motion Dataset (WOMD)~\cite{ettinger2021large} is limited to only three types: vehicle, pedestrian, cyclist.
In practice, we interact with agents who might be hard to fit into this taxonomy such as pedestrians on scooters or motor cyclists.
Moreover, the fidelity of the input features is quite limited to 3D boxes that hide many important details such as pedestrian postures and gaze directions.
2) Coverage of driving scene representation is centered around where the perception system detects objects.
The detection task becomes a bottleneck of transferring information to motion forecasting and planning when we are not sure if an object exist or not, especially in the first moments of an object surfacing.
We hope for more graceful transmission of information between the systems that is error-robust.
3) Training perception models to match these intermediate representations might evolve them into overly complicated systems that get evaluated on subtasks that are not well correlated with overall system quality.

The goal of this work is to provide a large-scale, diverse raw sensor dataset for the motion forecasting task.
\revonly{We aim to release a {\lidar} dataset with a similar format of WOD~\cite{sun2020scalability} for the motion forecasting task, with 100$\times$ more scenes than those available in WOD~\cite{sun2020scalability}.}
\arxivonly{We aim to augment WOMD~\cite{ettinger2021large} with {\lidar} data in a similar format of WOD~\cite{sun2020scalability} for the motion forecasting task, with 100$\times$ more scenes than those available in WOD~\cite{sun2020scalability}.} 
To the best of our knowledge, it is the largest publicly available {\lidar} dataset across perception or motion forecasting tasks (Table \ref{tab:stat_db_intro}).
To overcome the huge data storage problem and make the dataset user-friendly for academic research, we adopt state-of-the-art {\lidar} compression technology~\cite{zhou2022riddle}.
It reduces the {\lidar} dataset by $\sim8\times$, resulting in the final {\dbname} data to be around 2.3 TB.

To demonstrate the usefulness of the new {\lidar} data, we propose a novel and simple motion forecasting baseline, which leverages raw {\lidar} data to boost prediction accuracy.
Instead of jointly training the perception and prediction networks, which demands huge memory footprint, we take a two-stage approach: 
we first apply a perception model~\cite{sun2022swformer} to extract embedding features from {\lidar} data.
Then, during training, we feed these embeddings to a motion forecasting model, {\wf}~\cite{nayakanti2022wayformer}.
We \revonly{collect machine-generated labels in the same format as those in WOMD~\cite{ettinger2021large}, and}evaluate the model with same metrics as WOMD~\cite{ettinger2021large}.
Experiments show that, with {\lidar} data, the {\wf} model has a 2\% mAP increase for Vehicle and Pedestrian prediction respectively.
This indicates that the {\dbname} brings useful information and can further improve motion forecasting models' performance.

\revonly{The {\dbname} data will be made publicly available to the research community, and we hope it will provide new directions and opportunities in developing end-to-end motion forecasting models.}
\arxivonly{The {\dbname} data has been made publicly available to the research community, and we hope it will provide new directions and opportunities in developing end-to-end motion forecasting models.}
Additionally, {\dbname} opens the door for new research on detection and tracking with a very large amount of 3D boxes and tracks.

We summarize the contributions of our work as follows:
\begin{itemize}
    \item We release the largest scale {\lidar} dataset for motion forecasting 
    with high quality raw sensor data across a wide spectrum of diverse scenes.
    \item We provide a baseline that boosts the motion forecasting performance using the raw data, demonstrating the efficacy of the sensor inputs.
    \item We design an encoding scheme that utilizes intermediate perception representations as a feature extraction utility for motion forecasting models.
\end{itemize}

\section{Related Work}

\parab{Motion forecasting datasets.} 
There has been an increasing number of motion forecasting 
datasets released~\cite{ettinger2021large,Woven2020dataset,Woven2020dataset_corl,caesar2020nuscenes,Argoverse2,zhan2019interaction,robicquet2016learning,coifman2017critical,pellegrini2009you,lerner2007crowds,benfold2011stable,breuer2020opendd,bock2020ind}. Table~\ref{tab:stat_db_intro} shows the comparison for several most relevant motion forecasting datasets which aim at real-world urban driving environments.
The Woven Planet prediction dataset~\cite{Woven2020dataset} processed raw data through their perception system with over 1000 hours of logs for the traffic agents. 
nuScenes~\cite{caesar2020nuscenes} is an autonomous driving dataset that supports detection, tracking, prediction and localization.
But both of these~\cite{Woven2020dataset,caesar2020nuscenes} did not explicitly collect or upsample diverse, complex or interactive driving scenarios.
Argoverse~\cite{argoverse1,Argoverse2} mined for vehicles in various scenarios (\emph{e.g.} intersections, dense traffic).
The INTERACTION dataset~\cite{zhan2019interaction} collects some interactive scenarios (\emph{e.g.}, roundabouts, ramp merging).
The Shifts~\cite{malinin2021shifts} dataset targets vehicle motion prediction and has the longest duration. 
However, many of these long-duration datasets \cite{Woven2020dataset,zhan2019interaction,malinin2021shifts} lack {\lidar} data, blocking the exploration of end-to-end motion forecasting.
nuPlan~\cite{caesar2021nuplan}, an ego vehicle's planning dataset, released only a subset of the  {\lidar} sequences. 
Compared with other autonomous driving perception datasets~\cite{sun2020scalability, caesar2020nuscenes,geiger2013vision,behley2019semantickitti} that provide {\lidar} frames,  {\dbname} is significantly larger in terms of the total time, number of scenes and object interactions.

\parab{Motion forecasting modeling.} 
A popular approach is to render each input frame as a rasterized top-down image where each channel represents different scene elements~\cite{chai2019multipath,cui2019multimodal,lee2017desire,hong2019rules,casas2018intentnet,zhao2021tnt}. 
Another method is to encode agent state history using temporal modeling techniques like RNN~\cite{mercat2020multi,khandelwal2020if,alahi2016social,rhinehart2019precog} or temporal convolution~\cite{liang2020learning}.
In these two methods, relationships between each entity are aggregated through pooling~\cite{zhao2021tnt,ye2022dcms,alahi2016social,gupta2018social,lee2017desire,mercat2020multi}, soft attention~\cite{mercat2020multi,zhao2021tnt} and graph neural networks~\cite{casas2020spagnn,khandelwal2020if,liang2020learning}.
Recently, some work~\cite{nayakanti2022wayformer,shi2022motion} explore the Transformer~\cite{vaswani2017attention} encoder-decoder structure for multimodal motion prediction.
We choose {\wf}~\cite{nayakanti2022wayformer} as our motion forecasting baseline: it is a state-of-the-art model, which can flexibly integrate features from our new {\lidar} modality.

\parab{{\lidar} data compression.}
Releasing the {\lidar} data for our dataset presents a data storage challenge: without {\lidar} compression techniques, the raw sensor data of {\dbname} exceeds 20 TB.
As valuable as the data is, the size is inconvenient for fast distribution in the research community.
Fortunately, in recent years, there is a growing interest in the {\lidar} point cloud compression techniques.
For example, one major stream of work, octree-based methods, which represent and compress quantized point clouds~\cite{botsch2002efficient,schnabel2006octree}, has been released as a point cloud compression standard~\cite{graziosi2020overview}.
More recently, neural network based octrees squeeze methods have been proposed, such as Octsqueeze~\cite{huang2020octsqueeze}, MuSCLE~\cite{biswas2020muscle} and VoxelContextNet~\cite{que2021voxelcontext}.
Alternatively, {\lidar} point clouds can be stored as range images.
A family of image-based compression methods have been adapted for the task.
For example,  traditional methods such as JPEG, PNG and TIFF have been applied to compressing range images~\cite{ahn2014large,houshiar20153d}.
Recently, RIDDLE~\cite{zhou2022riddle} extends such method by applying a deep neural network and delta encoding to compress range images.
We adopt the delta encoder of RIDDLE~\cite{zhou2022riddle} and reduce the raw sensor data by $\sim8\times$.

\section{Dataset}

In this section, we describe the {\dbname} dataset  statistics, the {\lidar} data format, and the compression technique used to reduce the storage footprint.

\begin{figure}[t]
\begin{center}
\includegraphics[width=1.0\linewidth]{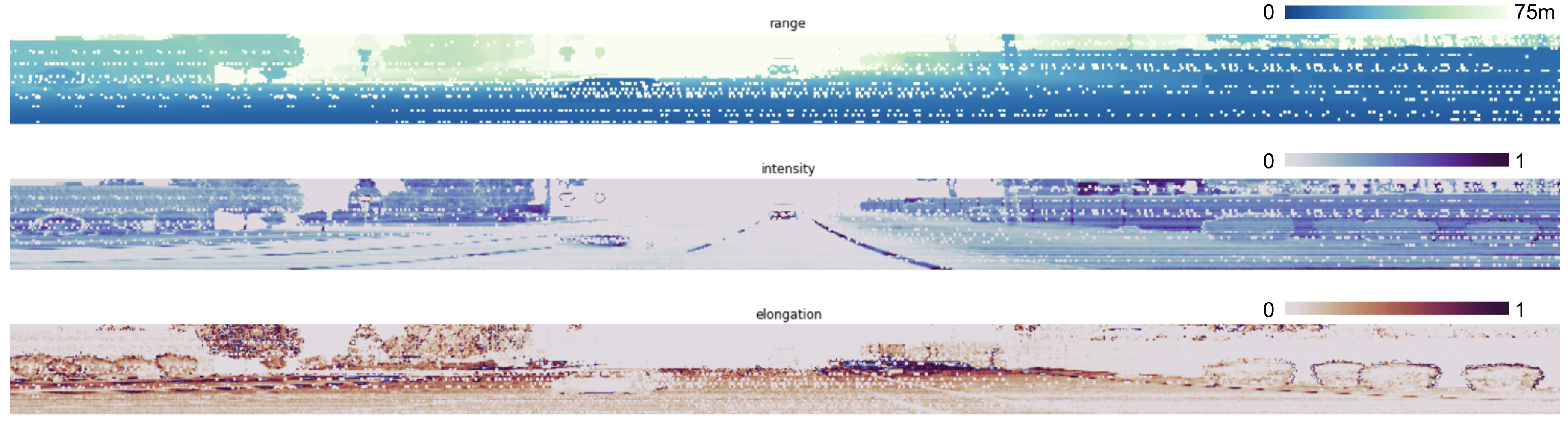}
\end{center}
  \caption{Visualization of a range image from the top {\lidar} sensor in {\dbname}. The three rows are showing range, (normalized) intensity, and (normalized) elongation from the first LiDAR return (second return omitted due to brevity). We crop the range images to only show the front 180$^{\circ}$.} 
\vspace{-3mm}
\label{fig:range_img_viz}
\end{figure}
\subsection{Dataset Statistics}\label{sec:data_stat}
\revonly{To evaluate motion forecasting models on {\dbname}, we collect labels generated by an offboard perception system in the same format as {WOMD}~\cite{ettinger2021large}.}
\arxivonly{To evaluate motion forecasting models, we leverage existing labels gathered from {WOMD}~\cite{ettinger2021large}.}
We follow the WOMD dataset format, and extract 9 second scenarios containing {\lidar} data.
\revonly{Same as WOMD, we split the data into a 70\% training, 15\% validation, and 15\% test set.}
\arxivonly{{\dbname} is split into a 70\% training, 15\% validation, and 15\% test set with the same run segments in WOMD.}
For training a motion forecasting model, it is sufficient to only use the past and current timestamps' {\lidar} data, while the future timestamps are used as ground truth to calculate loss and metrics.
We only release the first 1 second {\lidar} data for each scene.
This helps reduce the 87.9\% size of the raw {\lidar} data.
However, it still reaches $\sim$20TB data storage. We further apply a {\lidar} compression method to reduce its size (Section~\ref{sec:lidar_compression}).

\parab{Datasets comparison}: Compared with WOD~\cite{sun2020scalability}, one of the largest datasets for the perception task, {\dbname} contains \textbf{100$\times$} more scenes, \textbf{80$\times$} total hours. nuScenes~\cite{caesar2020nuscenes} is currently the only other {\lidar} dataset suitable for the motion forecasting task.
{\dbname} is significantly larger than nuScenes, with 104k \textbf{(100$\times$)} segments  and 574 hours \textbf{(100$\times$)} of total time (see Table~\ref{tab:stat_db_intro}).

\subsection{{\lidar} Data Format}
{\lidar} data is encoded in {\dbname} as range images $\in \mathbb{R}^{h\times w\times 6}$.
Following the format of WOD~\cite{sun2020scalability}, the first two returns of {\lidar} pulse are provided.
Range images are collected from five {\lidar} sensors.
For top {\lidar}, $h=64, w=2650$. For other sensors, $h=116, w=150$.
Each pixel in the range images includes the following:
\begin{itemize}
    \item Range (scalar): The distance between the origin of {\lidar} sensor frame and the {\lidar} point.
    \item Intensity (scalar): It is a measurement describing the return strength of the laser pulse that produces the {\lidar} point, which is partially based on the reflectivity of the object struck by the laser pulse.
    \item Elongation (scalar): The elongation of the laser pulse beyond its normal width.
    \item Vehicle pose ($\in \mathbb{R}^3$): The pose of the vehicle when the {\lidar} point is captured.
\end{itemize}
The range image format is necessary to exploit efficient compression schemes to reduce storage requirements (Section \ref{sec:lidar_compression}).
\figref{fig:range_img_viz} shows the different features that constitute the range images through mono-chromatic images, one for each feature.
We provide a tutorial\footnote{\revonly{The anonymous URL where it will be provided.}\arxivonly{\url{https://bit.ly/tutorial-womd-lidar}}} to show how to decompress range images and convert them into the features above.

\begin{figure*}[t]
\begin{center}
\includegraphics[width=1.0\linewidth]{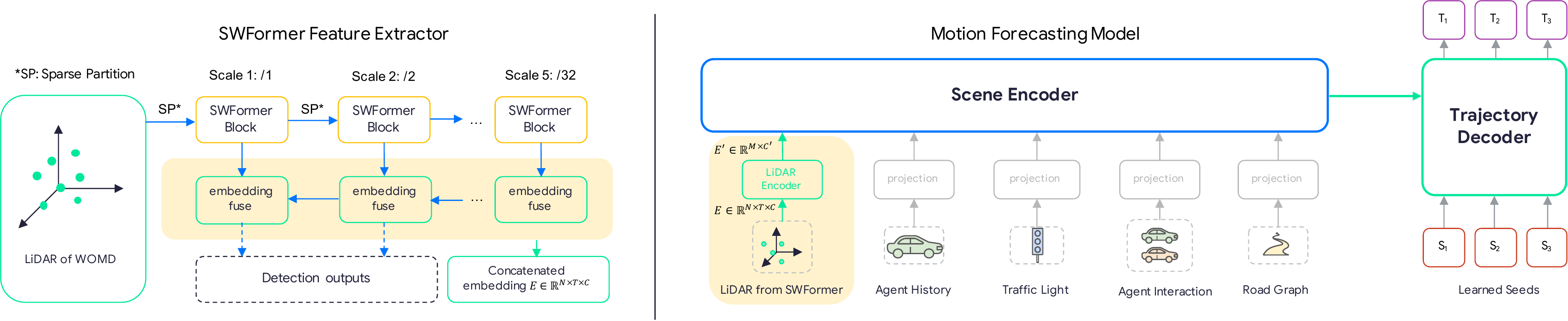}
\end{center}
   \caption{Model structures of {\lidar} encoder (left) and motion forecasting model (right). To encode {\lidar} data, we adopt a pre-trained {\swf}~\cite{sun2022swformer} model and extract the embedding features (which can be decoded to produce detection results). Those features (in the light yellow box) from different scales are concatenated and fed to a {\wf}~\cite{nayakanti2022wayformer} model as a new modality feature for the motion forecasting task.}
\vspace{-3mm}
\label{fig:model_struct}
\end{figure*}

\subsection{{\lidar} Data Compression}\label{sec:lidar_compression}

Storing raw sensory data is prohibitively expensive.
Therefore, we apply the delta encoding compressor proposed in~\cite{zhou2022riddle}.
We use a non-deep-learning version of the algorithm for fast compression and decompression.
This compression is lossless  under a pre-specified quantization precision.
Therefore, we do not expect to impact end-to-end learning.

The basic idea of the algorithm is to use a previous  pixel value in the range image to predict the next valid pixel (the closest valid one on its right in the spatial domain). Instead of storing the absolute pixel values, we store the residuals between the predictions and the original pixel values.
Since the residuals have a more concentrated distribution (especially on quantized range images) with lower entropy, they are compressed to a much smaller size with \texttt{varint} coding followed by zlib compression.

In our implementation, we quantize the range image channels with the following precision: range 0.005m, intensity 0.01m, elongation 0.01m, pose translation 0.0001m, pose rotation 0.001 radians.
We leverage the default \texttt{varint} coding from the publicly available Google Protobuf implementation (for \texttt{uint} and \texttt{bool} fields).
We will release our compression algorithm together with the dataset.

\section {Motion Forecasting Model with {\lidar}}

To validate the effectiveness of {\dbname}, we train a {\wf}~\cite{nayakanti2022wayformer} model using {\lidar} embeddings as a baseline. 
We describe the details of the motion forecasting model and the {\lidar} encoder (\figref{fig:model_struct}) in this section.

\subsection{Motion Forecasting Model}
We extend the {\wf}~\cite{nayakanti2022wayformer} model to incorporate raw {\lidar} data.
It adopts a transformer based scene encoder which is flexible to plug in features from various modalities.
The transformer fuses features from agent history states, traffic light signals, agent interaction states and road graph features, We add additional {\lidar} modality fed to the scene encoder.
The features of {\lidar} modality are generated from a {\swf}~\cite{sun2022swformer} extractor and a {\lidar} encoder.
During the training, we freeze the gradients of the {\swf} feature extractor and update only the {\lidar} encoder's model parameters.
After applying the scene encoder to fuse multi-modal features, the output embeddings are fed to the trajectory decoder to produce the final predicted trajectories.

\subsection{{\lidar} Encoding Scheme}
\label{sec:lidar_enc}
We adopt a pre-trained {\swf}~\cite{sun2022swformer} to extract {\lidar} embeddings.
The {\swf} is trained on WOD~\cite{sun2020scalability} for the 3D object detection task.
The {\swf} adopts sparse partition operators and transformer based layers to encode {\lidar} data from different scales.
We extract the embedding features which are used to produce detection results in the detection heads as the input to the scene encoder of {\wf} model.
These features effectively encode rich information of objects and context environment from noisy {\lidar} points.
To provide context agent information, we lower the detection confidence threshold to produce more but less reliable detected objects.
This increases the recall of the detection results but decreases the precision.
In addition to the embedding features, we also pad more features:
\begin{itemize}
    \item Detected box coordinates: We append the detected boxes center coordinates to emphasize the potential detected objects positions.
    \item Detected box size: The height, width, length of the boxes provide hints of objects from different categories.
    \item Foreground probability from the segmentation head: This helps reduce the noise from detection results. 
\end{itemize}
The output tensor $E$ from {\swf} with padded features is a $N\times T \times C$ tensor, where $N$ is the number of detected boxes, $T$ is the number of input frames, $C$ is the feature size.
To adapt $E$ to be compatible as input for the scene encoder of {\wf}, we flatten the first two dimensions as the token dimension.
A one-layer Axial Transformer~\cite{ho2019axial} is applied as a {\lidar} encoder to project the output tensor $E$ to be a fixed $M$-token tensor $E^\prime\in\mathbb{R}^{M\times C^\prime}$ with the same feature size as other modalities.

\section{Experiments}


\begin{table*}[t]
\begin{center}
\small
\begin{tabular}{@{}cc|ccc|ccc|ccc@{}}
\toprule
\multirow{2}{*}{Set} &
  \multirow{2}{*}{Model} &
  \multicolumn{3}{c|}{Vehicle} &
  \multicolumn{3}{c|}{Pedestrian} &
  \multicolumn{3}{c}{Cyclist} \\
 &
   &
  minADE $\downarrow$ &
  MR $\downarrow$ &
  mAP $\uparrow$ &
  minADE $\downarrow$ &
  MR $\downarrow$ &
  mAP $\uparrow$ &
  minADE $\downarrow$ &
  MR $\downarrow$ &
  mAP $\uparrow$ \\ \midrule
\multicolumn{1}{l}{} &
  LSTM~\cite{ettinger2021large} &
  \multicolumn{1}{c}{1.34} & 
  \multicolumn{1}{c}{0.25} & 
  \multicolumn{1}{c|}{0.23} & 
  \multicolumn{1}{c}{0.63} &
  \multicolumn{1}{c}{0.13} &
  \multicolumn{1}{c|}{0.23} &
  \multicolumn{1}{c}{1.26} &
  \multicolumn{1}{c}{0.29} &
  \multicolumn{1}{c}{0.21} \\ \cmidrule(l){2-11} 
\begin{tabular}[c]{@{}c@{}}Standard\\ Validation\end{tabular} &
  Wayformer~\cite{nayakanti2022wayformer}
   & 1.10
   & 0.18
   & 0.35
   & 0.54
   & 0.11
   & 0.35
   & 1.08
   & 0.22
   & \textbf{0.29}
   \\ \cmidrule(l){2-11} 
 &
  \begin{tabular}[c]{@{}c@{}}Wayformer\\ + LiDAR\end{tabular} 
   & \textbf{1.09}
   & \textbf{0.17}
   & \textbf{0.37}
   & \textbf{0.54}
   & \textbf{0.10}
   & \textbf{0.37}
   & \textbf{1.06}
   & \textbf{0.21}
   & 0.28
   \\ \bottomrule
\end{tabular}
\end{center}
\caption{\txbu{Marginal} \textbf{metrics on the} \txbu{standard} \textbf{validation set}. All metrics computed at 8s. 
\revonly{We implement and compare our own WOMD~\cite{ettinger2021large} baseline, {\wf}~\cite{nayakanti2022wayformer} and {\wf} trained with {\lidar} data on the {\dbname} standard motion forecasting track.}
\arxivonly{We compare baseline {\wf}~\cite{nayakanti2022wayformer} and {\wf} trained with {\lidar} data on the {\dbname} standard motion forecasting track.}
}
\vspace{-3mm}
\label{tab:exp_res_standard}
\end{table*}

\subsection{Experiment Setup}\label{sec: exp_setup}
\parab{{\lidar} Feature Extractor.} We train the {\swf}~\cite{sun2022swformer} on WOD~\cite{sun2020scalability} as the {\lidar} feature extractor.
We set batch size as 4, training 80,000 steps on 64 V3 TPUs.
The IOU thresholds for vehicles and pedestrians are 0.7 and 0.5 respectively.
In the original {\swf} inference stage, the boxes are filtered if the predicted confidence is less than 0.5.
To extract feature embeddings, we need more context information and high recall of the detection results.
Thus, we lower the box confidence threshold $\tau$ to be 0.1 (see ablation study in Section~\ref{sec:ablation}).
The extracted embeddings are 128D vectors.
With box coordinates ($x$, $y$, $z$), box size (width, length, height) and foreground probability, the final {\lidar} features are 135D vectors ($C=135$) fed to the scene encoder of the {\wf}~\cite{nayakanti2022wayformer}.
We set the maximum number of detected boxes in each frame as 140 ($N \leq 140)$.
If there are more than 140 detected objects, we discard the detected objects with low box confidence scores.
We set the number of output tokens of the {\lidar} encoder as $M=10$ before sending the embeddings to the scene encoder.

\parab{Motion Forecasting Model.} We 
use a batch size of 16 and train the {\wf} model with 1.2M steps on 16 V3 TPUs.
We project all modalities to the same feature size of 256D ($C^\prime=256$), then utilize cross-attention with latent queries to reduce the number of tokens to 192.
The scene encoder has 2 transformer layers. 
{\wf} encodes the history states of 1 second (10 steps at 10Hz) and predicts K=6 trajectories for each agent's future 8 seconds.

\subsection{Metrics}
Given an input sample, a motion forecasting model predicts $K$ trajectories for $N$ agents in the scene for the future $T$ steps $\mathbf{x}^k = \{x_{i, t}\}_{i=1:N, t=1:T}$.
We denote the corresponding ground truth trajectories as $\mathbf{y} = \{y_{i, t}\}_{i=1:N, t=1:T}$. We inherit the WOMD motion forecasting challenge metrics~\cite{ettinger2021large}.

\parab{minADE.} 
The minimum Average Displacement Error calculates the $\ell_2$ distance between the predicted trajectory which is closest to the ground truth across all time steps:
\begin{equation}\label{eq:minADE}
    \text{minADE} = \min_{k}\frac{1}{NT}\sum_i\sum_t||\mathbf{x}^k_{i, t} - \mathbf{y}_{i, t}||_2
\end{equation}

\parab{Miss Rate (MR).} MR measures whether the closest predicted trajectory $\min_{k}\mathbf{x}^k_{i, t}$ matches the ground truth $\mathbf{y}_{i, t}$. The MR at time step $t$ is calculated as:
\begin{equation}\label{eq:mr_at_t}
    \text{MR}_t = \min_k\vee_i\neg\texttt{IsMatch}(\mathbf{x}^k_{i, t}, \mathbf{y}_{i, t})
\end{equation}
More details of the function \texttt{IsMatch} implementation can be found in the WOMD dataset~\cite{ettinger2021large}.

\parab{Mean Average Precision (mAP).} mAP is similar to the one for object detection task~\cite{lin2014microsoft}.
It computes precision-recall curve's integral area by varying confidence threshold for the predicted trajectories.
The criteria of judging whether a trajectory is a true positive, false positive, \emph{etc}. is consistent with the MR definition in Eq.~\ref{eq:mr_at_t}.
For each object, only the trajectory with the highest confidence is used to calculate the mAP for the corresponding true positive.

\subsection{Baseline Model Performance}

We evaluate our baseline model on the {\dbname} validation set.
The results are shown in Table~\ref{tab:exp_res_standard}. 
With {\lidar} features, our model performs better than {\wf} for vehicle, pedestrians and cyclists on the Missing Rate (MR) metric, with 0.01 decrease in each category respectively.
This indicates {\lidar} information provides location hints for {\wf}.
For minADE metrics, the results are roughly the same.
{\wf} with {\lidar} inputs also achieves 2\% increase in mAP for vehicle and pedestrian categories. 
This is because {\lidar} features provide more information about the object locations, shapes and interactions with other objects. 
They help the {\wf} model understand the scene and predict more accurate trajectories.
For cyclists, there is a minor regression in mAP.
It is likely due to the fact that the {\lidar} points are noisy in this category and we may need a better encoding method to extract useful information.

\subsection{Ablation Study}
\label{sec:ablation}
In the following experiments, we report the average minADE, MR and mAP across vehicle, pedestrian and cyclist categories at 3s, 5s and 8s on the validation set.

\parab{Threshold of {\swf} to extract embeddings.}
As described in Sec.~\ref{sec: exp_setup}, we lower the {\swf} threshold $\tau$ to get high recall of detected boxes so that we could get more context information in the scene.
We sweep the threshold of {\swf} from 0.0 to 0.5 (default value of {\swf}), and generate different training datasets extracted from {\dbname} and evaluate the corresponding performance of the baseline model.
When the threshold $\tau$ is lower, the number of predicted boxes from {\swf} becomes larger. 
This brings more context information for motion forecasting model while it also brings more noise in the inputs.
As shown in Table~\ref{tab:exp_threshold_embed}, the {\wf}'s performance is not so sensitive to $\tau$.
When $\tau=0.1$, the {\wf} with {\lidar} inputs achieves the best performance.
When $\tau$ further increases, the number of detected boxes becomes smaller and may result in loss of useful information.

\begin{table}[t]
\begin{center}
\small
\begin{tabular}{@{}ccccc@{}}
\toprule
\multicolumn{1}{c|}{Threshold $\tau$} & minADE $\downarrow$ & MR $\downarrow$ & mAP $\uparrow$  \\ \midrule
\multicolumn{1}{c|}{0.0} & 0.5692  & 0.1401  & 0.4005  \\
\multicolumn{1}{c|}{0.1} & \textbf{0.5553} & \textbf{0.1292} & \textbf{0.4191}  \\
\multicolumn{1}{c|}{0.3} & 0.5623 & 0.1399 & 0.4102  \\
\multicolumn{1}{c|}{0.5} & 0.5675 & 0.1410 & 0.4087  \\ \bottomrule
\end{tabular}
\end{center}
\caption{Experiment results of sweeping {\swf} threshold $\tau$ to extract embeddings. The metrics are evaluated on {\dbname} validation set, averaged across categories, and over results at 3s, 5s, and 8s.
}
\vspace{-3mm}
\label{tab:exp_threshold_embed}
\end{table}
\begin{table}[t]
\begin{center}
\small
\begin{tabular}{@{}ccccc@{}}
\toprule
\multicolumn{1}{c|}{Model} & minADE $\downarrow$ & MR $\downarrow$ & mAP $\uparrow$  \\ \midrule
\multicolumn{1}{c|}{No boxes coordinates} & 0.5852  & 0.1594  & 0.3947  \\
\multicolumn{1}{c|}{No boxes sizes} & 0.5773 & 0.1476 & 0.4008  \\
\multicolumn{1}{c|}{No foreground prob.} & 0.5601 & 0.1331 & 0.4110  \\
\multicolumn{1}{c|}{Wayformer with LiDAR} & \textbf{0.5553} & \textbf{0.1292} & \textbf{0.4191}   \\ \bottomrule
\end{tabular}
\end{center}
\caption{Experiment results of masking out additional features in {\lidar} encoding (Sec.~\ref{sec:lidar_enc}). The metrics are evaluated on {\dbname} validation set, averaged across categories, and over results at 3s, 5s, and 8s.}
\vspace{-3mm}
\label{tab:exp_feat_mask}
\end{table}

\begin{figure*}[t]
\begin{center}
\includegraphics[width=0.95\linewidth]{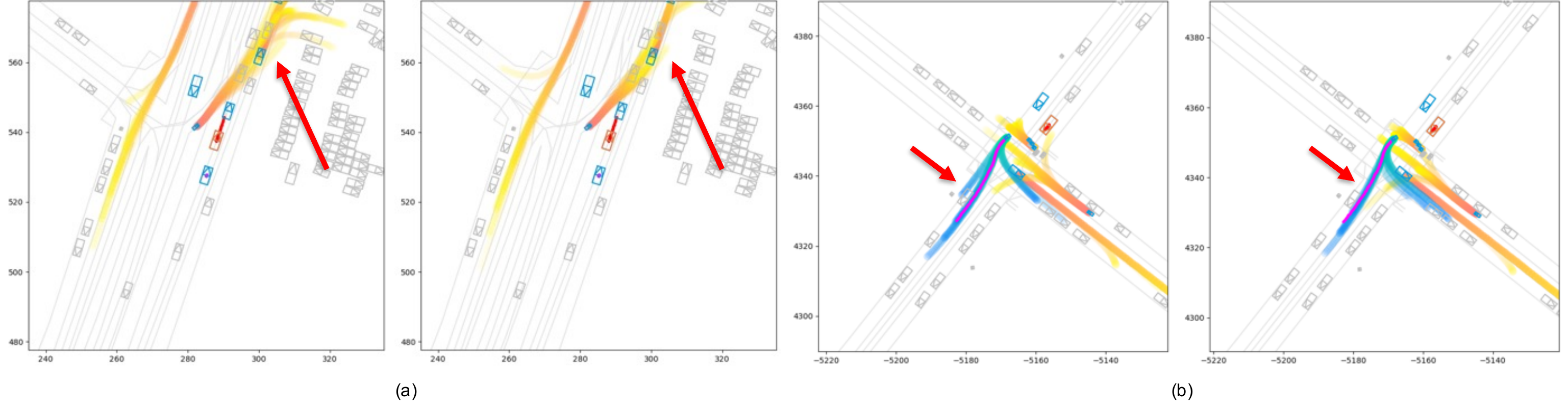}
\end{center}
\vspace{-3mm}
   \caption{Visualization of prediction result comparison between {\wf}~\cite{nayakanti2022wayformer} (sub-figures on the left) and {\wf} with {\lidar} inputs (sub-figures on the right). Fig (a): With {\lidar} information the predicted trajectories avoid crashing into parked cars. Fig (b): The predicted trajectories of cyclists avoid crashing into cars.
   Legends in the figure: Yellow and blue trajectories are predictions for different agents, while blue trajectories are highlighted ones. Red dotted lines are labeled ground truth trajectories for agents in the scene.
   \revonly{More visualization results are available in the supplementary material.}}
   \vspace{-3mm}
\label{fig:exp1_viz}
\end{figure*}

\parab{Different embedding features.} There are three additional features (Sec.~\ref{sec:lidar_enc}) included in the embedding output from the {\lidar} encoder: detected box coordinates, size and foreground probability. 
We mask out each feature and check the {\wf} model performance in Table~\ref{tab:exp_feat_mask}.
The experiments show that without box coordinates, the minADE, MR, mAP regress by 0.0299, 0.0302, 0.0244 respectively.
This indicates that aside from the {\swf} embedding features, the box coordinates play an important role in motion forecasting .
Compared to masking out box coordinates, masking out box sizes has a smaller regression, with minADE, MR increased by 0.022 and 0.0184 and mAP decreased by 0.0183. 
Foreground probability also contributes slightly to the overall performance, with regression in the minADE, MR, mAP as 0.0048, 0.0039, 0.0081 respectively.

\begin{table}[t]
\begin{center}
\small
\begin{tabular}{@{}ccccc@{}}
\toprule
\multicolumn{1}{c|}{\# tokens of embeddings} & minADE $\downarrow$ & MR $\downarrow$ & mAP $\uparrow$  \\ \midrule
\multicolumn{1}{c|}{16} & 0.6011  & 0.1702  & 0.3811  \\
\multicolumn{1}{c|}{32} & 0.5888 & 0.1610 & 0.3907  \\
\multicolumn{1}{c|}{64} & 0.5797 & 0.1503 & 0.3998  \\
\multicolumn{1}{c|}{192} & \textbf{0.5553} & \textbf{0.1292} & \textbf{0.4191}   \\ \midrule
\multicolumn{1}{c|}{\# layers of transformer} & minADE $\downarrow$ & MR $\downarrow$ & mAP $\uparrow$  \\ \midrule
\multicolumn{1}{c|}{1} & 0.5711  & 0.1440  & 0.3991  \\
\multicolumn{1}{c|}{2} & \textbf{0.5553} & \textbf{0.1292} & \textbf{0.4191}  \\
\multicolumn{1}{c|}{3} & 0.5561 & 0.1325 & 0.4112   \\ \bottomrule
\end{tabular}
\end{center}
\caption{Experiment results of scene encoder's \#tokens and \#transformer layers. The metrics are evaluated on {\dbname} validation set, averaged across categories, and over results at 3s, 5s, and 8s.}
\vspace{-3mm}
\label{tab:scene_encoder}
\end{table}

\parab{{\wf} modeling.} We study the {\wf} hyper-parameters in motion forecasting.
Specifically, we conduct experiments to investigate the impact of number of tokens and layers of the scene encoder. 
This is because the scene encoder provides encoded embeddings for the trajectory decoder in the prediction stage.
The embedding quality plays an important role for the motion forecasting task.
As shown in Table~\ref{tab:scene_encoder}, the number of embedding tokens impacts quality more than the number of scene encoder transformer layers.
When we increase the token size from 16 to 192 (the default {\wf} setting), the minADE and MR decrease from 0.6011 to 0.5553 and 0.1702 to 0.1292, respectively, and mAP increases from 0.3811 to 0.4191. This indicates that when the token size increases, more information will be encoded in the embeddings for motion prediction.

We also vary the number of transformer blocks from 1 to 3 (Table~\ref{tab:scene_encoder}). The performance of {\wf} model first improves (\# layers increases from 1 to 2) and then regresses (\# layers increases from 2 to 3). 
Thus, we set the optimal value of \# layers of the scene encoder as 2.
\subsection{Qualitative Results}
We visualize the {\wf} prediction results on {\dbname} to check the quality motion forecasting.
\textbf{Please check the supplementary video for more visualization results.}


\noindent\textbf{Visualization of {\wf} prediction results.}
We visualize some prediction results and conduct analysis on the prediction quality.
As shown in \figref{fig:exp1_viz}, with {\lidar} inputs, {\wf} model avoids collision into vehicles, pedestrians and cyclists.
Specifically, in Fig~\ref{fig:exp1_viz}(a), with {\lidar} information, the predicted trajectories avoid crashing into parked cars.
In Fig~\ref{fig:exp1_viz}(b), the predicted trajectories of cyclists avoid crashing into cars.
We observe more reasonable predicted trajectories, matching the improved performance in Table~\ref{tab:exp_res_standard}.

\section {Conclusion and Future Directions}
\parab{Conclusion.}
\revonly{In this work, we release {\dbname}, the largest scale {\lidar} dataset in the community, containing {\lidar} point clouds for more than 100,000 scenes.}
\arxivonly{In this work, we augment WOMD with the largest scale {\lidar} dataset in the community, containing {\lidar} point clouds for more than 100,000 scenes.}
To resolve the huge data storage requirements, we adopt state-of-the-art {\lidar} data compression technology and successfully reduce the dataset size to be less than 2.5 TB.
To evaluate the suitability of {\lidar} to the motion forecasting task, we provide a {\wf} baseline trained with {\lidar}\revonly{ and machine generated labels in the same format as WOMD~\cite{ettinger2021large}}. 
Experiments show that {\lidar} data brings improvement in the motion forecasting task.

\parab{Limitations and future work.}
1) In this work, we only trained {\wf} and {\wf} + {\lidar} models. We will investigate end-to-end models that can directly encode {\lidar} point clouds with motion forecasting task in mind.
2) The {\swf} detector, which serves as the point cloud encoder in our model, can only represents object-level information. We will look into some approaches that can leverage scene-level information, that are not sensitive to the detection prediction thresholds.
3) Another interesting direction is to explore methods that solely depends on the sensor data to avoid the dependency on  human-defined object interface.




{
\bibliographystyle{ieee_fullname}
\bibliography{icra24.bbl}
}


\newpage
\section*{APPENDIX}

\section{Supplementary Dataset Details}
\begin{table*}[b]
\begin{center}
\small
\begin{tabular}{c | *{6}{c}}
\toprule
& INTERACTION  & Woven Planet & Shifts  & Argoverse 2 & nuScenes & {\dbname} \\
\midrule
Offboard Perception  & \cmark &  & & & & \cmark \\
Mined for Interestingness & - & - & - & \cmark & - & \cmark \\
Traffic Signal States  & & \cmark & \cmark & & & \cmark \\
\bottomrule
\end{tabular}
\end{center}
\caption{Comparison of the popular behavior prediction and motion forecasting datasets. 
``-'' indicates that the data is not available or not applicable. ``Offboard perception'' is checked if the labels were auto-labeled by offboard perception which can generate high-quality labels. ``Mined for Interestingness'' is checked if the dataset mined interesting interactions after the data collection. ``Traffic Signal States'' is checked if the dataset provided traffic light states.
}
\label{tab:supp_dataset_comparison}
\end{table*}

\begin{table*}[b]
\begin{center}
\small
\begin{tabular}{@{}c|ccc|c|ccc@{}}
\toprule
\# output tokens ($M$) & minADE $\downarrow$ & MR $\downarrow$ & mAP $\uparrow$ & \# layers & minADE $\downarrow$ & MR $\downarrow$ & mAP $\uparrow$ \\ \midrule
5  & 0.5700 & 0.1501 & 0.3999 & 1 &  \textbf{0.5553} & \textbf{0.1292} & \textbf{0.4191}  \\
10 & \textbf{0.5553} & \textbf{0.1292} & \textbf{0.4191}  & 2 & 0.5613 & 0.1392 & 0.3998  \\
20 & 0.5594 & 0.1313 & 0.4102 & 3 & 0.5610 & 0.1398 & 0.4001 \\ \bottomrule
\end{tabular}
\end{center}
\caption{Experiment results of the number of output tokens $M$ and the number of transformer layers in the {\lidar} encoder. The metrics are evaluated on {\dbname} validation set, averaged across categories, and over results at 3s, 5s, and 8s.
}
\label{tab:supp_lidar_encoder}
\end{table*}

\parab{Dataset comparison.} We provide more details of the dataset comparison in Table~\ref{tab:supp_dataset_comparison}.
In Table~\ref{tab:stat_db_intro} of the main paper, ``Sampling Rate'' is the data collection rate in Hz.
``3D Maps'' indicates whether the dataset provided the 3D Map information.
``Dataset Size'' entries were collected by Argoverse 2~\cite{Argoverse2}.
Combining Table~\ref{tab:supp_dataset_comparison} and Table~\ref{tab:stat_db_intro} of the main paper, we provide the complete comparison between our {\dbname} and other datasets.

\parab{Supplementary Details of {\dbname}.} Map data is encoded as a set of polylines and polygons created from curves sampled at a resolution of 0.5 meters following~\cite{ettinger2021large,gao2020vectornet}.
Traffic signal states is also provided along with other static map feature types (\emph{e.g.}, lane boundary lines, road edges and stop signs).
We followed~\cite{ettinger2021large} to mine the interesting scenarios in our {\dbname}.

\begin{figure*}[!t]
\centering

\includegraphics[width=0.3\linewidth]{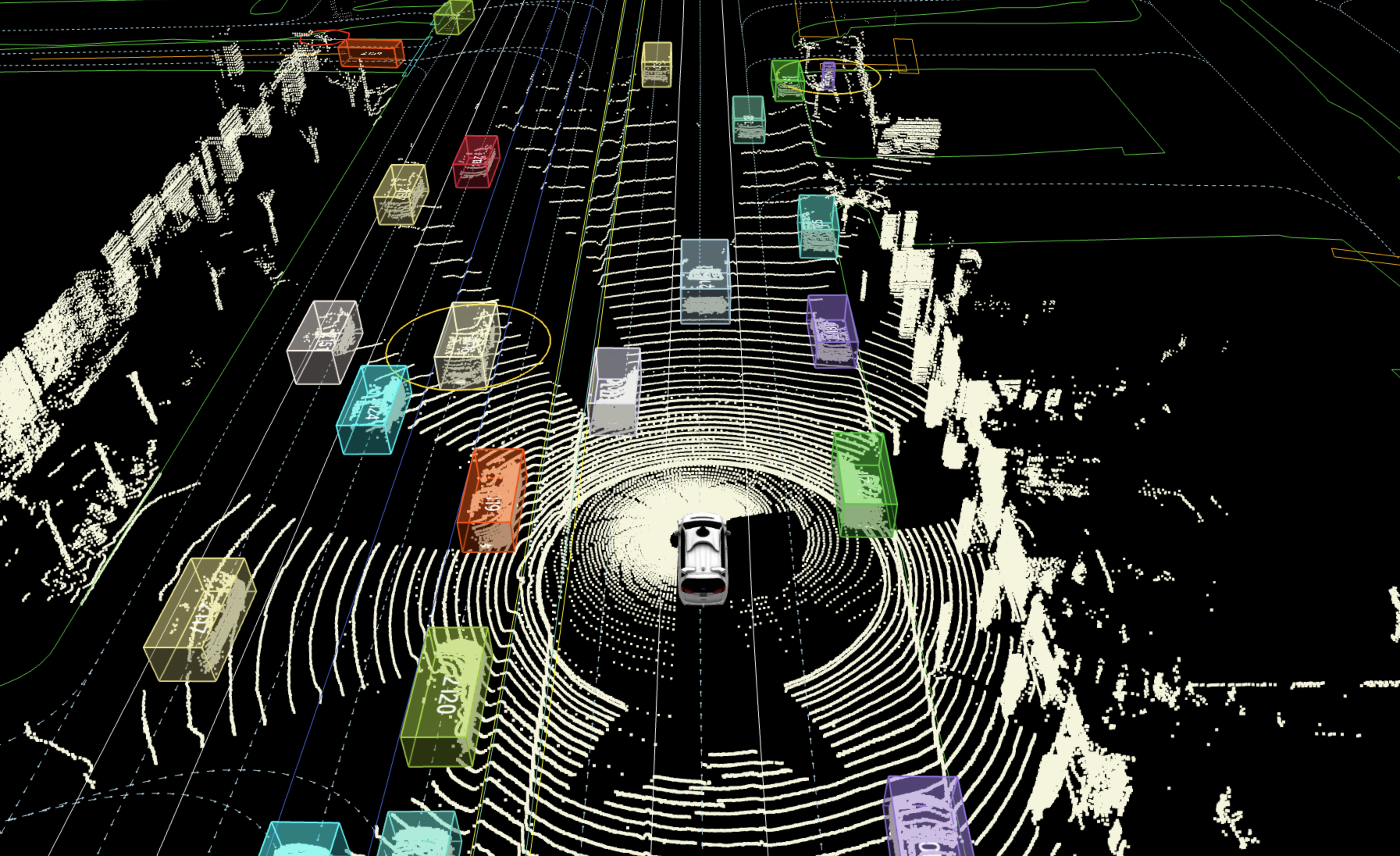}
\includegraphics[width=0.3\linewidth]{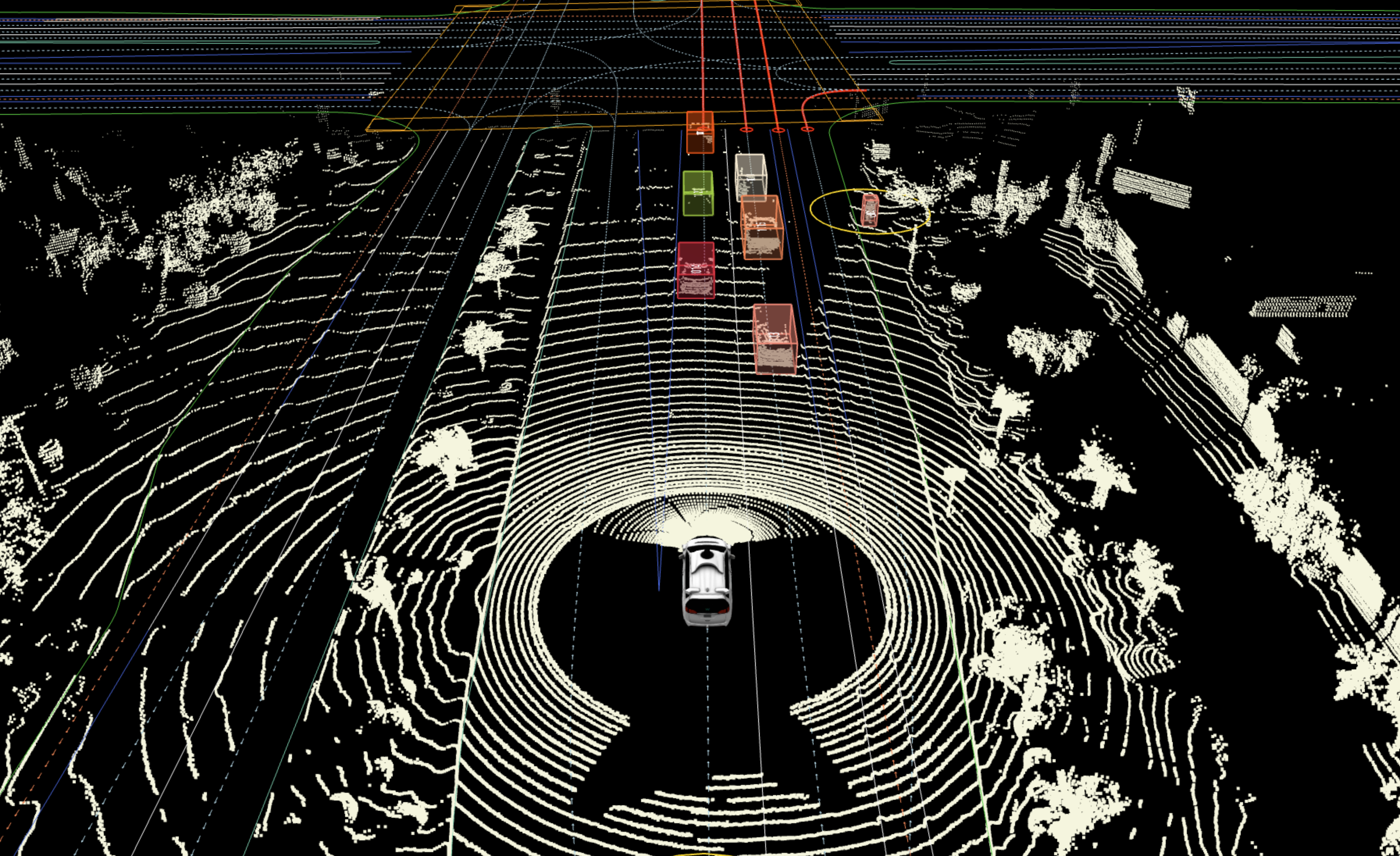}
\includegraphics[width=0.3\linewidth]{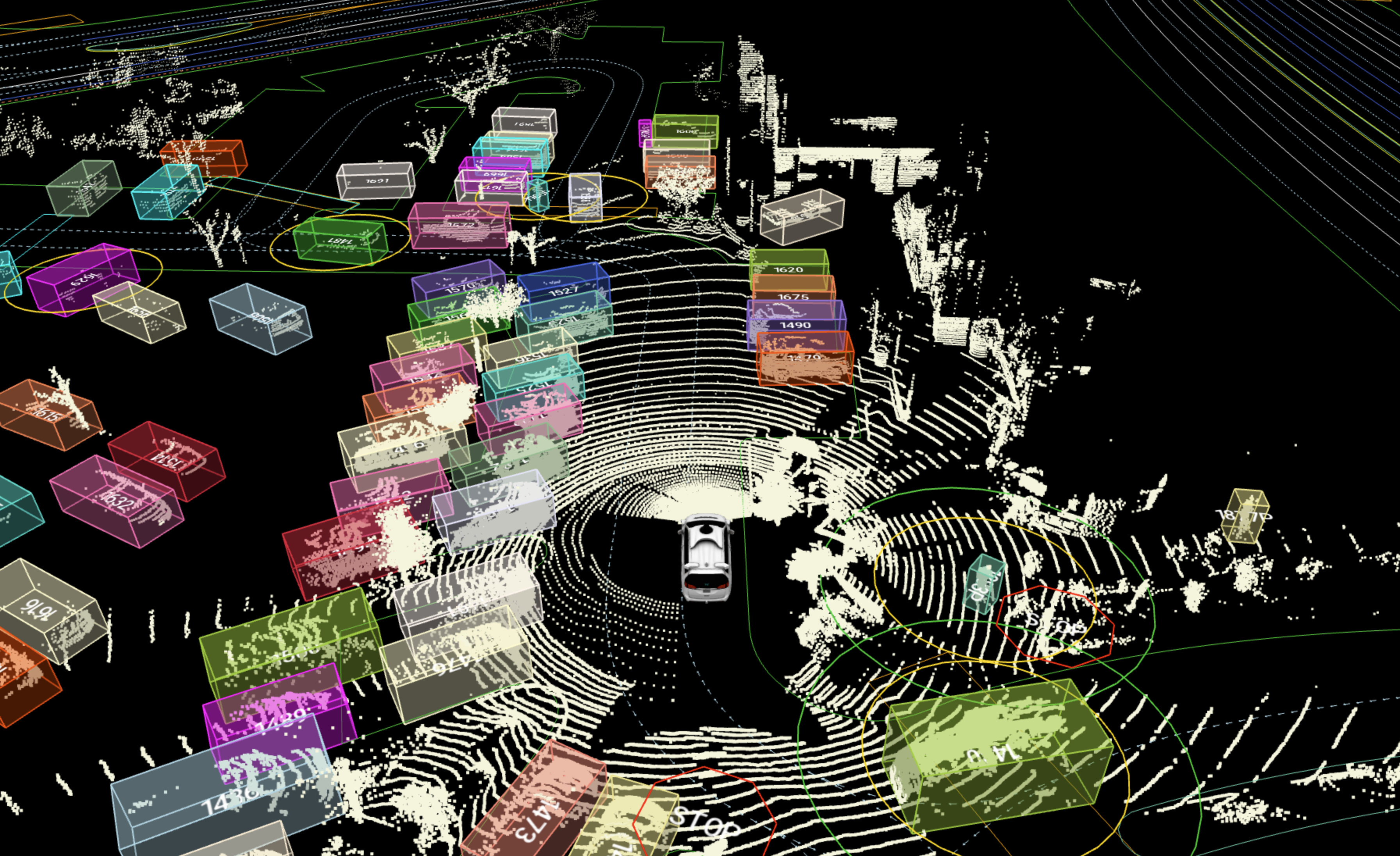}

\caption{Scenario visualizations with {\lidar}. Better viewed in color and zoom in for more details.}
\label{fig:supp_viz}
\vspace{-6mm}
\end{figure*}

\section{Visualization}

\parab{Scenario Videos with {\lidar}.} In \figref{fig:supp_viz}, we provide more visualization of scenarios with not only the bounding boxes of the agents of interest, but also the released high quality well calibrated {\lidar} data.

We provide some simulated scenes with both {\lidar} and labeled boxes on {\dbname}.
They are formulated as \texttt{mov} files in the supplementary materials.
Each video clip contains 11 frames in slow motion, with {\lidar} data visualized with the boxes of agents.
This is because we only release the first 11 frames' {\lidar} data in {\dbname}.

\begin{figure*}[!t]
\begin{center}
\includegraphics[width=1.02\linewidth]{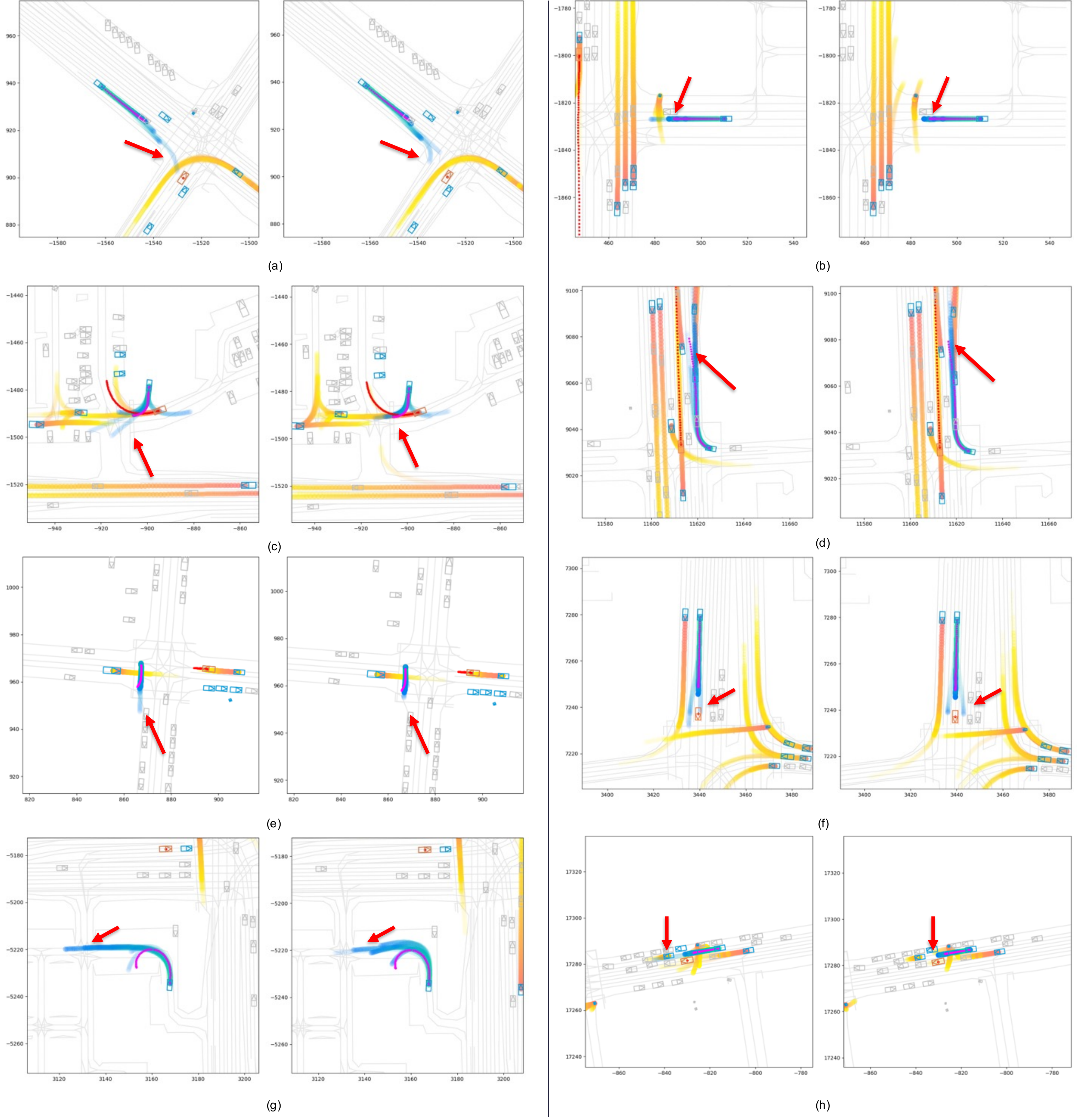}
\end{center}
\vspace{-3mm}
   \caption{Visualization of prediction result comparison between {\wf}~\cite{nayakanti2022wayformer} (sub-figures on the left) and {\wf} with {\lidar} inputs (sub-figures on the right).
   Legends in the figure: Yellow and blue trajectories are predictions for different agents, while blue trajectories are highlighted ones. Red dotted lines are labeled ground truth trajectories for agents in the scene. More visualization results are available in the supplementary material. Better viewed in color and zoom in for more details.}
\label{fig:supp_exp_viz}
\end{figure*}

\parab{{\wf} + {\lidar} prediction visualization.} We provide more visualization results in \figref{fig:supp_exp_viz}.
From the visualization results, our {\wf}~\cite{nayakanti2022wayformer} + {\lidar} model tries to avoid collision into other agents (vehicles, pedestrians and cyclists) in the motion forecasting task. 
This is consistent with the improved performance in Table~\ref{tab:exp_res_standard}. 

\section{Experiments}

\parab{Ablation Study of {\lidar} Encoder.}
We provide the ablation study of {\lidar} Encoder described in Section 4.2 in the submission.
Specifically, we study the number of output tokens $M$ and the number of transformer layers. 
Experiment results are shown in the Table~\ref{tab:supp_lidar_encoder}.

From the Table~\ref{tab:supp_lidar_encoder}, we find when $M$ increases, the final performance of {\wf} first increases and then decreases.
We set the optimal value of $M$ as 10 in our experiments.
On the other side, {\lidar} encoder is not so sensitive to the number of transformer layers.
There is a slight regression when the number of layers increases.
To achieve best performance and fast training speed, we set the number of layers as 1 in our experiments.

\revonly{\section{Discussion}

There are some more potential interesting directions and contributions by releasing {\dbname}.
We would like to provide more insights here:
\begin{itemize}
    \item {\dbname} enables various {\lidar} detection and tracking tasks, with 100$\times$ labeled {\lidar} dataset than WOD~\cite{sun2020scalability} (though the labels are machine generated by an offline perception system).
    \item We introduce a new benchmark on {\dbname} as the first baseline showing the utility of the {\lidar} on the motion forecasting task. 
    We believe there is a lot more headroom for the motion forecasting task, and releasing {\dbname} opens such opportunity to realize it.
\end{itemize}
}

\newpage


\end{document}